\def\year{2021}\relax
\documentclass[letterpaper]{article} 
\usepackage{aaai21}  
\usepackage{times}  
\usepackage{helvet} 
\usepackage{courier}  
\usepackage[hyphens]{url}  
\usepackage{graphicx} 
\usepackage{natbib}  
\usepackage{caption} 
\usepackage{cleveref} 
\usepackage{booktabs} 
\usepackage{multirow} 
\title{Deep learning-based fast solver of the shallow water equations}
\author{
    Mojtaba Forghani\textsuperscript{\rm 1}, Yizhou Qian\textsuperscript{\rm 2},
    Jonghyun Lee\textsuperscript{\rm 3},
    Matthew W. Farthing\textsuperscript{\rm 4}, Tyler Hesser\textsuperscript{\rm 4}, Peter K\@. Kitanidis\textsuperscript{\rm 2,5}, and Eric F\@. Darve\textsuperscript{\rm 1,2}
    \\
}
\affiliations{
    \textsuperscript{\rm 1}Department of Mechanical Engineering, Stanford University, CA\\
    \textsuperscript{\rm 2}Institute for Computational and Mathematical Engineering, Stanford University, CA\\
    \textsuperscript{\rm 3}Department of Civil and Environmental Engineering, University of Hawaii at Manoa, Honolulu, HI\\
    \textsuperscript{\rm 4}U.S\@. Army Engineer Research and Development Center, Vicksburg, MS\\
    \textsuperscript{\rm 5}Department of Civil and Environmental Engineering, Stanford University, CA\\
}

\begin{document}

\maketitle

\begin{abstract}
Fast and reliable prediction of river flow velocities is important in many applications, including flood risk management. The shallow water equations (SWEs) are commonly used for this purpose. However, traditional numerical solvers of the SWEs are computationally expensive and require high-resolution riverbed profile measurement (bathymetry). In this work, we propose a two-stage process in which, first, using the principal component geostatistical approach (PCGA) we estimate the probability density function of the bathymetry from flow velocity measurements, and then use machine learning (ML) algorithms to obtain a fast solver for the SWEs. The fast solver uses realizations from the posterior bathymetry distribution and takes as input the prescribed range of BCs. The first stage allows us to predict flow velocities without direct measurement of the bathymetry. Furthermore, we augment the bathymetry posterior distribution to a more general class of distributions before providing them as inputs to ML algorithm in the second stage. This allows the solver to incorporate future direct  bathymetry measurements into the flow velocity prediction for improved accuracy, even if the bathymetry changes over time compared to its original indirect estimation. We propose and benchmark three different solvers, referred to as PCA-DNN (principal component analysis-deep neural network), SE (supervised encoder), and SVE (supervised variational encoder), and validate them on the Savannah river, Augusta, GA. Our results show that the fast solvers are capable of predicting flow velocities for different bathymetry and BCs with good accuracy, at a computational cost that is significantly lower than the cost of solving the full boundary value problem with traditional methods.
\end{abstract}

\section{Introduction}
\label{intro}

Estimation of riverine flow velocities is important in many practical applications such as the study of river morphodynamics, safe and efficient maritime transportation, and flood risk management~\cite{overdeep, morph, Casas, Westaway, Lane}. In order to accurately estimate flow velocities under user specified boundary conditions (BCs), such as the discharge and the free-surface elevation, as well as the bathymetry, we require an accurate predictor of the flow velocities given the bathymetry and the BCs. The shallow water equations (SWEs) are typically used to solve this problem~\cite{Landon}. However, current numerical solvers of the SWEs are computationally expensive. This is a major shortcoming of these methods, since BCs in rivers can vary widely and thus having a ``fast online predictor'' of the flow velocities is very important, in particular, in situations when a range of conditions need to be evaluated quickly to address questions related to navigability or to asses the risk of flooding. Furthermore, these solvers typically require a fairly high resolution image of the bathymetry as simulation input. However, direct high-resolution bathymetric surveys~\cite{Casas} are time consuming and costly for long river reaches.

\begin{figure}[htbp]
\centering
\includegraphics[width=0.98\linewidth]{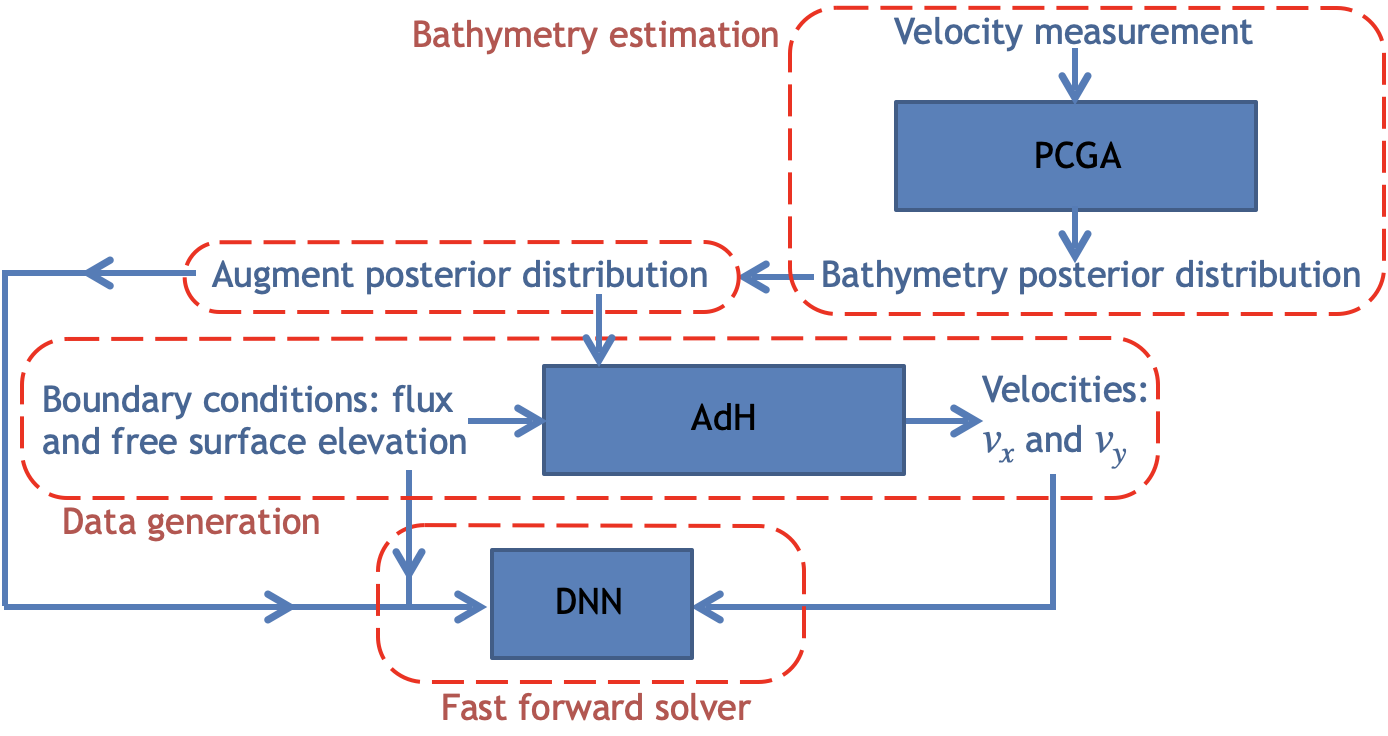}
\caption{The schematic of the development of the forward solver. First, we estimate the posterior distribution of the bathymetry via PCGA, then augment this distribution to a more general distribution and use AdH to generate velocities. Finally, the DNNs are trained with these data, which will be used as fast forward solvers.}
\label{sketch}
\end{figure}

In this work, we propose a two-stage process in which, first, the river bathymetry at a site of interest is estimated using the principal component geostatistical approach (PCGA)~\cite{PCGA_Lee, PCGA_Kitanidis} from velocity measurements (thus addressing the issue of not having access to direct bathymetry measurement), and then the distribution of estimated bathymetry is augmented to a more general distribution and combined with different BCs to obtain a fast solver of the SWEs (thus addressing the high computational cost of numerical solvers). \Cref{sketch} shows the steps in the proposed approach schematically. Note that our solver is capable of taking either directly measured bathymetry or its estimated distribution as inputs. Thus, the purpose of the posterior augmentation stage is to allow our solver to include a more general class of bathymetries into their prediction capability, for instance, when new direct bathymetry measurements become available and they have changed over time compared to their original indirect estimation due to sediment deposition or erosion. 

\section{Methods}
\label{meth}

\begin{figure}[htbp]
\vspace{-0.1cm}
\centering
\includegraphics[width=1.0\linewidth]{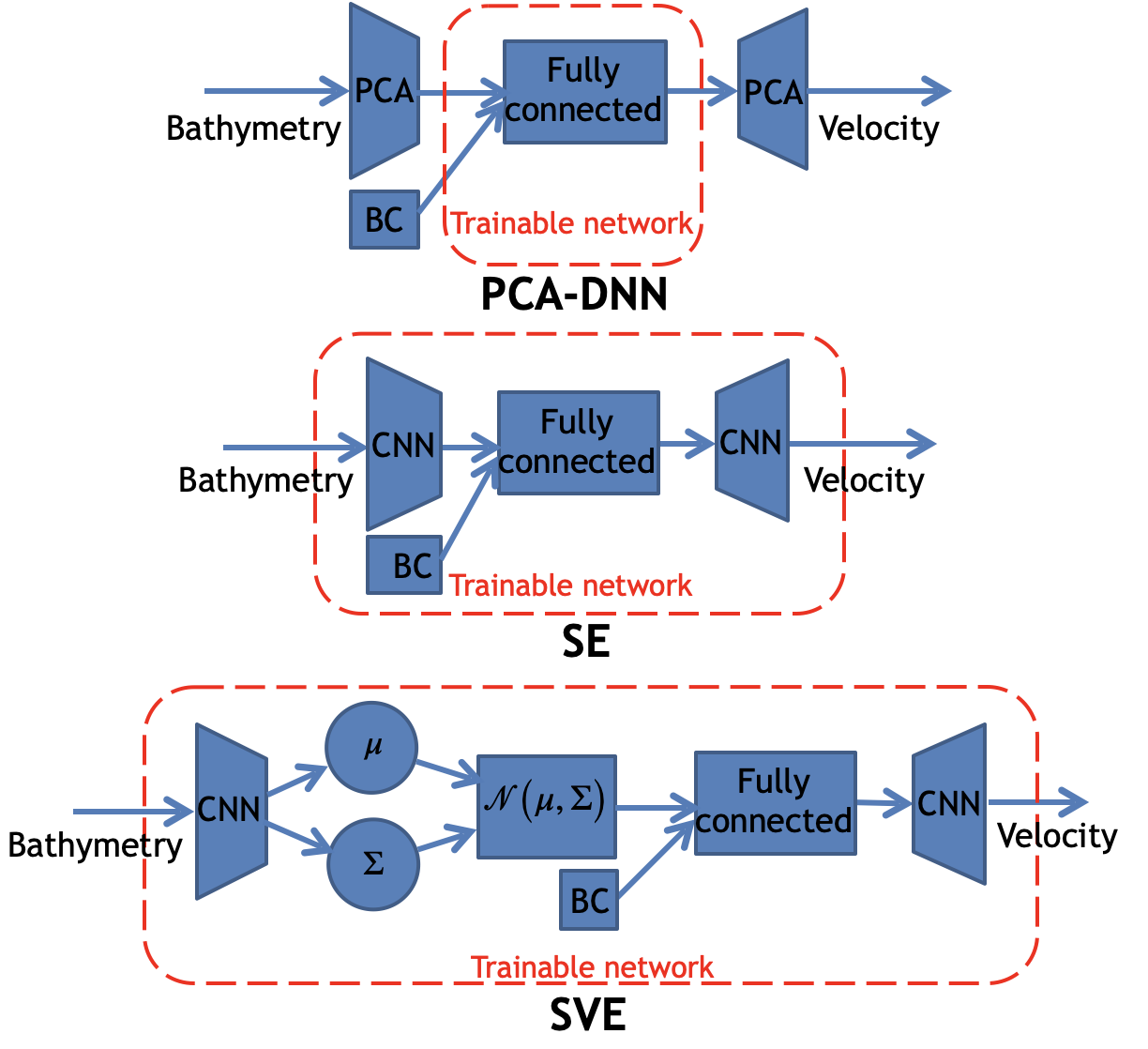}
\caption{Schematic of the PCA-DNN, SE, and SVE.}
\label{DNNs_sketch}
\end{figure}

In this work, we use three different deep learning methods as fast SWE solvers. These methods, shortly, are referred to as PCA-DNN (principal components analysis-deep neural network), SE (supervised encoder), and SVE (supervised variational encoder). The schematic of these methods are shown in \cref{DNNs_sketch}. The PCA-DNN method consists of first, a low-rank approximation of data via PCA-based linear projection, and then applying DNN to the reduced-dimension data~\cite{PCA_DNN_Hojat}. SE is similar to an autoencoder (AE)~\cite{Kramer}, except, it is used for supervised learning. In SE architectures, a high-dimensional input (bathymetry) is fed as the input to the network, where its dimension is reduced via a convolutional neural network (CNN), then it is combined with the BCs (with two elements: discharge and the free-surface elevation), passes through a fully connected network, and finally is augmented to the high dimensional output (the velocity) via another CNN. SVE is also similar to a variational autoencoder (VAE)~\cite{VAE_ref}, but it is used in supervised learning. The SVE has a similar structure as SE, except the middle layer which defines a random variable based on a multivariate normal distribution.

\section{Data preparation}
\label{data_prep}

The first step in the data preparation process is applying PCGA to flow velocity observations taken from the river in order to obtain an estimation of the bathymetry in the area of interest. In the following, we refer to this as the PCGA posterior distribution. Here, we have applied PCGA to the roughly one mile reach of the Savannah river, Augusta, GA. The flow velocity measurements in this section are generated synthetically via a numerical solver, referred to as Adaptive Hydraulics (AdH) SWEs module~\cite{AdH}, by first calculating the flow velocities corresponding to the reference bathymetry of the Savannah river and then applying Gaussian noise with a standard deviation equal to 10$\%$ of the largest simulated flow velocity, in order to ensure the synthetically generated flow velocities include the noise commonly observed in the field observations. Once the noisy synthetic velocity measurements are generated, we can use PCGA~\cite{PCGA_Lee} to obtain an estimation of the bathymetry. This estimation is in the form of a distribution (the posterior distribution).

While the PCGA posterior distribution provides a reasonable estimate of the uncertainty associated with the currently available dataset, we also consider an additional augmentation of the training data in order to broaden the range of bathymetries for which the proposed forward solvers are valid. For instance, when new direct bathymetry measurement becomes available and it has changed over time. To perform the augmentation, the synthetic data that are fed to the DNN architectures are generated by adding a Gaussian kernel of the following form to the PCGA estimation:
\begin{equation}
\textnormal{cov}(x,y)= \beta^2 \exp\left( -\frac{\Delta x^2}{l_x^2}-\frac{\Delta y^2}{l_y^2} \right)
\label{cov_x_y}
\end{equation}
Here, $\beta= 1.2$ m, $l_x= 115$ m, and $l_y= 29$ m ($x$ is the along-river direction while $y$ is the across-river direction). We then add a scaling factor to generated bathymetries that shrinks the variations near the shore, in order to capture the fact that the variations of the generated bathymetries near the shore are generally smaller than in the middle of the river. We also generate BC samples, extracted from the United States Geological Survey (USGS) gauge data of Savannah river, and provide them, along with the bathymetries sampled from the augmented distribution, to AdH, in order to obtain the flow velocities. The bathymetry/BC/flow velocity datasets are fed to the DNNs to obtain forward solvers.

\section{Performance in the presence of full bathymetry measurement}
\label{global_full}

\Cref{error_global} summarizes the root mean square errors (RMSEs) in estimating flow velocity magnitudes using different methods, when full bathymetry measurements are provided as inputs. A total of 4,000 river profiles (dataset size) have been used as the training set, 500 profiles for the validation set, and 450 profiles for the test set. In order to have a fair comparison between different methods, we used the same latent space dimension of 50 in all methods \cite{SWE_2021}. The errors in \cref{error_global} for SVE and SE are significantly lower than PCA-DNN, indicating that the non-linear dimension reduction contained in SVE and SE is more accurate than a linear, PCA-based approach. \Cref{hype_global} summarizes the hyperparameters used in different solvers. The table shows the different parameter values used in our networks during the hyperparameter tuning along with the final chosen value, which had the best performance (shown in bold in the table).

\begin{table}[htbp]
    \centering
    \begin{tabular}{lllll}
        \toprule
        Error (RMSE [m/s]) & \multicolumn{4}{c}{Fast forward solver}\\
        \cmidrule{2-4} \cmidrule{5-5}
        {} & PCA-DNN & SE & SVE \\
        \midrule
        Train set & 0.0515 & {\bf 0.0269} & 0.0286\\
        Validation set & 0.0570 & {\bf 0.0374} & 0.0398\\
        Test set & 0.0546 & {\bf 0.0381} & 0.0398\\
        \bottomrule
    \end{tabular}
    \caption{Comparison between the error of different solvers when predicting the magnitude of the flow velocity.}
    \label{error_global}
\end{table}

\begin{table*}[htbp]
    \centering
    \begin{tabular}{p{0.27\textwidth}p{0.2\textwidth}p{0.2\textwidth}p{0.2\textwidth}p{0.22\textwidth}}
        \toprule
        \multirow{2}{*}{DNN hyperparameter} & \multicolumn{3}{c}{Fast forward solver}\\
        \cmidrule{2-4} \cmidrule{5-5}
        {} & PCA-DNN & SE & SVE \\
        \midrule
        Type of layers   & Fully connected & Convolutional & Convolutional\\
        Batch normalization   & \{yes, {\bf no}\} & \{yes, {\bf no}\} & \{yes, {\bf no}\}\\
        Number of hidden layers   & \{{\bf 1},2,3,4,5,6\} & \{4,{\bf 6}\} & \{4,{\bf 6}\}\\
        Data normalization  & \{{\bf yes}, no\} & \{{\bf yes}, no\} & \{{\bf yes}, no\}\\
        Act. func. (hidden layer)  & \{{\bf tanh}, ReLU\} & \{{\bf tanh}, ReLU\} & \{{\bf tanh}, ReLU\}\\
        Act. func. (output layer)  & \{{\bf linear}, Sigmoid\} & \{{\bf linear}, Sigmoid\} & \{{\bf linear}, Sigmoid\}\\
        Batch size & \{8,{\bf 32},256,full\} & \{8,{\bf 32},256,full\} & \{8,{\bf 32},256,full\}\\
        Learning rate & \{0.01,{\bf 0.001},$10^{-4}$\} & \{0.01,{\bf 0.001},$10^{-4}$\} & \{0.01,{\bf 0.001},$10^{-4}$\}\\
        Reg. coeff. (easting) & \{0,0.00001,0.0001, 0.001,{\bf 0.01},0.1,1\} & \{0,0.00001,{\bf 0.0001}, 0.001,0.01,0.1,1\} & \{0,0.00001,{\bf 0.0001}, 0.001,0.01,0.1,1\}\\
        Reg. coeff. (northing) & \{0,0.00001,0.0001, 0.001,{\bf 0.01},0.1,1\} & \{0,0.00001,{\bf 0.0001}, 0.001,0.01,0.1,1\} & \{0,0.00001,0.0001, {\bf 0.001},0.01,0.1,1\}\\
        \bottomrule
    \end{tabular}
    \caption{The hyperparameters used in different solvers. The parameters in bold are the final values used in networks with the best performances. Act. func. is the activation function and reg. coeff. is the regularization coefficient. }
    \label{hype_global}
\end{table*}

\begin{figure}[htbp]
\vspace{-0.1cm}
\centering
\includegraphics[width=1.0\linewidth]{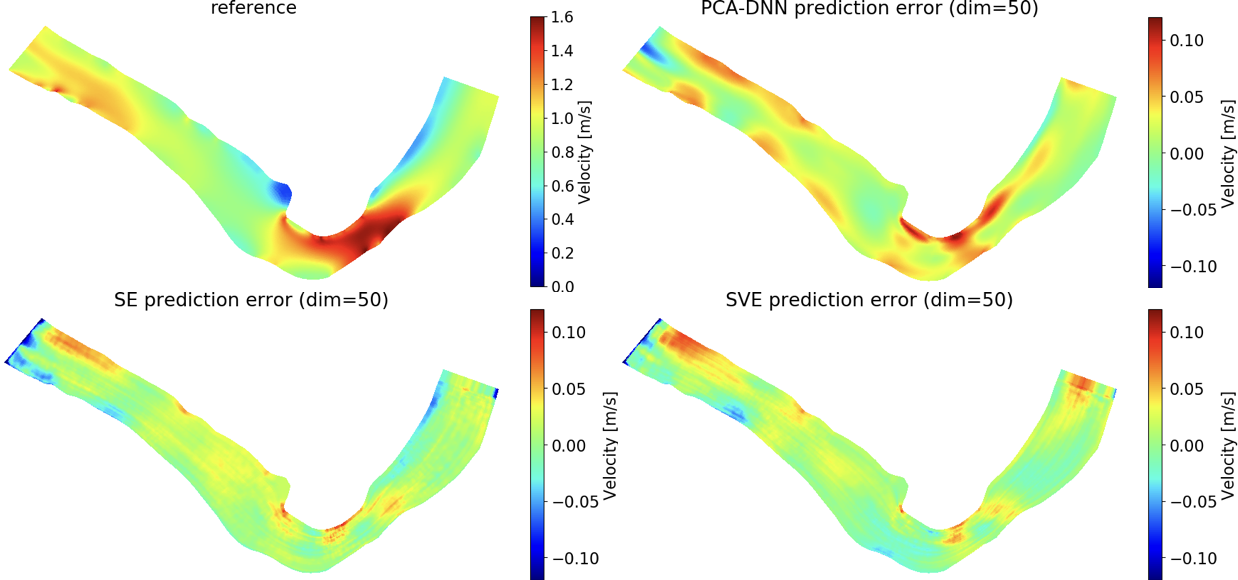}
\caption{Examples of the error in the prediction of the velocity magnitudes for different solvers for $z_f= 29.9$ m and $Q= 146.1$ m$^3$/s. SE and SVE outperform PCA-DNN.}
\label{plots_global_low}
\end{figure}

\Cref{plots_global_low} compares the performance of different methods when predicting the flow velocity magnitude of one of the members of the test dataset with BC values of free-surface elevation $z_f= 29.9$ m and discharge $Q= 146.1$ m$^3$/s. We observe that SE and SVE perform better than PCA-DNN, consistent with the result of \cref{error_global}. This could be due to the linear dimension reduction technique being used in this approach, which fails to capture non-linear features present in the data with 50 principal components (PCs).

\section{Performance in the presence of uncertain bathymetry}
\label{global_partial}

\begin{figure}[htbp]
\vspace{-0.1cm}
\centering
\includegraphics[width=1.0\linewidth]{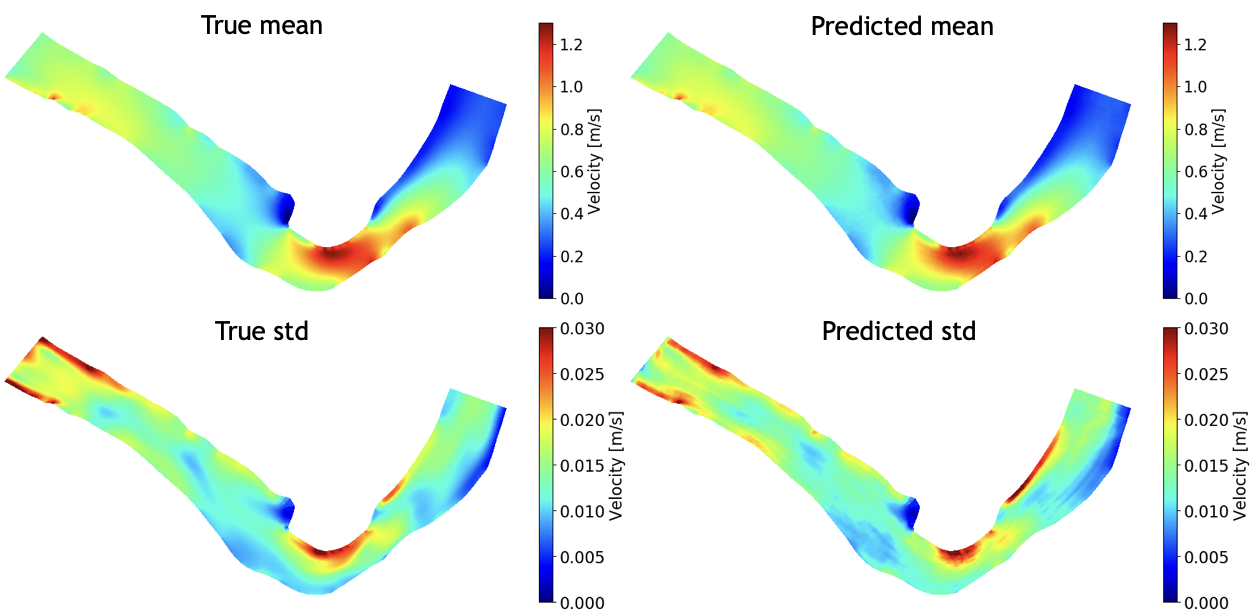}
\caption{Predicted mean and standard deviation of velocities for different solvers at $z_f= 33.9$ m and $Q= 651.2$ m$^3$/s. The ``reference" corresponds to the AdH prediction when bathymetries are generated from the PCGA posterior distribution. 
}
\label{plots_global_post}
\end{figure}

The results presented in the previous section provide informative evaluation metrics of different algorithms as forward solvers, that is, flow velocity predictors provided with bathymetry and BCs assuming the reference (true) bathymetries are known completely. In practice, however, there are many situations in which we do not have access to direct measurement of bathymetries, and all of our information must come from the solution of an inverse problem with an associated level of uncertainty. \Cref{plots_global_post} shows the reference mean and standard deviation of flow velocities in the easting direction obtained from AdH as well as the predicted mean and standard deviations obtained from the SE, respectively. The BCs for the simulations are $z_f= 33.9$ m and $Q= 651.2$ m$^3$/s. The results are based on first, generating 100 bathymetries directly from the PCGA posterior distribution, and then providing these profiles as inputs to either the AdH or any of the DNNs (with the given BCs); finally, the mean and standard deviation of their predicted velocities are calculated and plotted in \cref{plots_global_post}. 

We observe that the solver has been successful in finding the mean and the uncertainty. 
The great accuracy in \cref{plots_global_post} implies that even when indirect observations are available, we can use the same solvers, which are trained on the estimated bathymetry distribution from PCGA with augmentation, to predict the distribution of flow velocities as the BCs change. 

\section{Conclusion}
\label{conclusion}

In this work, we have presented a framework for fast prediction of the riverine flow velocities with user specified BCs and bathymetries. The training of all the presented methods can be performed on common personal computers without access to GPU and high-performance computing resources. More importantly, once the networks are trained, the predictions can be done in a few seconds, making online flow velocity estimations possible. Our results show that the computational efficiency is about {\bf three orders of magnitude faster} than standard SWE solvers such as AdH.

The combination of PCGA and our fast solvers provides a valuable tool that can be used even when the riverine bathymetry profiles are not a-priori available. That is, we do not need to measure riverbed profiles when training the network and designing the fast, reduced-order solver (offline stage). More importantly, even for the online prediction stage, we can predict distribution of flow velocities from the posterior distribution of the PCGA, without access to updated bathymetry observations. 

While all of the presented solvers are capable of providing reasonable prediction of the flow velocities, the better performance of SE and SVE methods implies that there are non-linear features present in the data that linear or partially-linear models such as PCA-DNN may not be able to capture accurately within the available computational limitations. 

\subsubsection{Acknowledgments.}
This research was supported by the U.S. Department of Energy, Office of Advanced Scientific Computing Research under the Collaboratory on Mathematics and Physics-Informed Learning Machines for Multiscale and Multiphysics Problems (PhILMs) project, PhILMS grant DE-SC0019453. This work was also supported by an appointment to the Faculty and Postdoctoral Fellow Research Participation Program at the U.S.\ Engineer Research and Development Center, Coastal and Hydraulics Laboratory administered by the Oak Ridge Institute for Science and Education through an inter-agency agreement between the U.S. Department of Energy and ERDC. The Chief of Engineers has granted permission for this publication.

\bibliography{bib.bib}

\end{document}